\documentclass[runningheads]{llncs}
\usepackage{graphicx}
\usepackage{amsmath}
\usepackage{amsfonts}
\usepackage{xcolor}
\usepackage{booktabs}
\usepackage{float}
\usepackage{subcaption}
\usepackage{bbm}
\usepackage{url}
\usepackage{hyperref}
\usepackage{wrapfig}

\begin{document}
\setlength{\abovedisplayskip}{10pt}
\setlength{\belowdisplayskip}{10pt}
\title{An Experimental Study on the Rashomon Effect of Balancing Methods in Imbalanced Classification}
\titlerunning{The Rashomon Effect of Balancing Methods}
%
\author{Mustafa Cavus\inst{1}\orcidID{0000-0002-6172-5449} \and
Przemysław~Biecek\inst{2,3}\orcidID{0000-0001-8423-1823}}
\authorrunning{Cavus and Biecek}
%
\institute{Eskisehir Technical University, Department of Statistics, Turkiye\\
\email{mustafacavus@eskisehir.edu.tr} 
\and Warsaw University of Technology, Faculty of Mathematics and Information Science, Poland \and University of Warsaw, Faculty of Mathematics, Informatics and
Mechanics, Poland}
\maketitle              

\begin{abstract}
Predictive models may generate biased predictions when classifying imbalanced datasets. This happens when the model favors the majority class, leading to low performance in accurately predicting the minority class. To address this issue, balancing or resampling methods are critical data-centric AI approaches in the modeling process to improve prediction performance. However, there have been debates and questions about the functionality of these methods in recent years. In particular, many candidate models may exhibit very similar predictive performance, called the Rashomon effect, in model selection, and they may even produce different predictions for the same observations. Selecting one of these models without considering the predictive multiplicity—which is the case of yielding conflicting models' predictions for any sample—can result in blind selection. In this paper, the impact of balancing methods on predictive multiplicity is examined using the Rashomon effect. It is crucial because the blind model selection in data-centric AI is risky from a set of approximately equally accurate models. This may lead to severe problems in model selection, validation, and explanation. To tackle this matter, we conducted real dataset experiments to observe the impact of balancing methods on predictive multiplicity through the Rashomon effect by using a newly proposed metric \textit{obscurity} in addition to the existing ones \textit{ambiguity} and \textit{discrepancy}. Our findings showed that balancing methods inflate the predictive multiplicity and yield varying results. To monitor the trade-off between the prediction performance and predictive multiplicity for conducting the modeling process responsibly, we proposed using the extended version of the performance-gain plot when balancing the training data.

\end{abstract}
\section{Introduction}

One of the most common challenges in classification tasks is class imbalance. In this case, models tend to produce biased predictions toward the majority class, resulting in significantly lower prediction performance for the minority class \cite{Khoshgoftaar_et_al_2007}. This issue, which can be solved using a model-centric approach, e.g., cost-sensitive methods, is solved much more practically with the data-centric AI approaches called data augmentation or data balancing methods that balance the majority and minority classes \cite{Zha_et_al_2023,Wang_et_al_2023,Singh_2023}. Many papers have been published to implement data balancing techniques for solving imbalance classification problems. A systematic study of these techniques can be found in \cite{Vargas_et_al_2023}. While the literature offers many balancing methods, none is universally superior \cite{Moniz_and_Monteiro_2021}. Each method has advantages and disadvantages, as oversampling-based methods may result in the model learning excessively, potentially leading to overfitting. Conversely, undersampling-based methods may result in information loss. Thus, the most suitable one should be chosen based on the task and the dataset's characteristics. 

The bias may be amplified throughout the processes of data-centric AI, depending on how the data is prepared. Similarly, the pre-processing of the data can also lead to changes in the model, such as changes in the model parameters \cite{Stando_et_al_2023}, the correlations between variables \cite{Patil_et_al_2020}, the importance of variables \cite{Alarab_and_Prakoonwit_2022}, model calibration \cite{Goorbergh_et_al_2022,Carriero_et_al_2024}, and the model fairness \cite{Simson_et_al_2024}. As a component of data pre-processing, balancing methods may affect the model behavior and, therefore, model selection. The strategy followed can lead to model changes, thereby diversifying the resulting set of models \cite{Watson_et_al_2023}. Despite various perspectives addressing these methods related to performance, only a few studies have delved into their impact on model behavior. It is shown that the SMOTE method effectively addressed imbalance issues while maintaining the original correlations between variables in the model \cite{Patil_et_al_2020}. In contrast, these methods altered feature importance in experiments conducted on two real datasets using various SMOTE variants \cite{Alarab_and_Prakoonwit_2022}. Stando et al. \cite{Stando_et_al_2023} conducted experiments on 21 datasets to investigate the impact of balancing methods on model behavior, and they showed the significant effects of balancing methods through partial dependence plots. They suggested that while utilizing balancing methods, it is essential to consider both performance gain and model behavior change. It is discussed that the full balancing is unnecessary to achieve optimal results and proposed partial resampling as the imbalanced ratio of 1.25 \cite{Kamalov_et_al_2022}. 

No studies have been conducted on the Rashomon effect of balancing methods on the model behavior, except \cite{watson_et_al_2023} which examines predictive multiplicity as it relates to the majority-minority structure of a dataset, showing that the minority group is particularly prone to predictive multiplicity. The Rashomon effect is a phenomenon wherein various models can model a dataset approximately equally accurately \cite{Breiman_2001}. Although this provides many alternatives in model selection, these models may yield conflicting predictions for any sample, which results in predictive multiplicity \cite{Marx_et_al_2020}. Thus, the potential negative consequences of randomly selecting a model must be considered as a part of responsible machine learning, as it masks the understanding of model differences. It is crucial because the performance of a model is not enough, and we need to explore model behavior in the phase of model selection \cite{Biecek_et_al_2023}. Hence, it is imperative to carefully consider the implications of the Rashomon effect in the model selection process. Since the critical effect of pre-processing steps in model selection, we consider the effect of balancing methods on model behavior, which is examined in the Rashomon set. To the best of our knowledge, this is the first paper related to the Rashomon effect of balancing methods on model behavior in terms of multiplicity. Moreover, we propose a new metric called \textit{obscurity} to measure the multiplicity of the Rashomon set in addition to existing metrics.

We focus on the following research questions to investigate the effect of balancing methods on model behavior in the data-centric AI view: \textbf{RQ1.} How do the balancing methods affect the predictive multiplicity of the models in the Rashomon set?, \textbf{RQ2.} How do the balancing methods affect the variable importance order discrepancy of the models in the Rashomon set? \textbf{RQ3.} Can the partial resampling be a solution against the model behavior change? and \textbf{RQ4.} Can the extended version of the performance gain plot be a solution to monitor the trade-off between performance gain and multiplicity? In the remainder of this paper, we first provide some preliminaries about the Rashomon effect and predictive multiplicity in Sec~\ref{sec:preliminaries}. We present the experiments in Sec~\ref{sec:experiments} and discuss the results in Sec~\ref{sec:results} and conclusions in the last section.

\section{Preliminaries}
\label{sec:preliminaries}

Let $\mathbf{D} = \{ \textbf{x}_i, y_i \}_{i=1}^n$ be a dataset of $n$ observations from $p$ variables where $\textbf{x}_i = [x_{i1}, x_{i2}, ..., x_{ip}] \in \textbf{X}$ and $y_i \in Y$ is the response vector. Also, let $F = \{f \mid f \colon \mathbf{X} \to Y\}$ be the space of all predictive models called \textbf{Hypothesis Space}. Furthermore, let $L \colon F \to \mathbb{R}$ denote the model's loss function. The goal is to find $f \in F$ that minimizes the expected value of the loss function $L$: 
\begin{equation}
    f = \operatorname*{argmin}_{f \in F} \mathbb{E}[L(f)].
\end{equation}

\noindent Note that expected loss $\mathbb{E}[L]$ is approximated with empirical loss calculated on data $\textbf{X}$. \textbf{Reference Model.} Let a model with the minimum loss function that we have found from $\hat{F}$, be called the reference model, and we shall denote it as $f_R$. In other words
\begin{equation}
f_R = \operatorname*{argmin}_{f \in F} \mathbb{E}[L(f)].
\end{equation}

\noindent \textbf{Rashomon Set.} In a learning problem, for a given loss function $L$, a reference model $f_R$, and the \textbf{Rashomon parameter} $\varepsilon > 0$ which limits the set, the Rashomon set $R_{L,\varepsilon}(f_R)$ is defined as:
\begin{equation}
R_{L,\varepsilon}(f_R) = \{f \in F \mid \mathbb{E}[L(f)] \leq \mathbb{E}[L(f_R)] + \varepsilon\}.
\end{equation}

\noindent It is not possible to access all possible models in $F$, so we are interested in an empirical hypothesis space $\hat{F}$, where $\hat{F} \subset F$, and an empirical Rashomon set, which we will refer to simply as the Rashomon set:
\begin{equation}
\hat{R}_{L,\varepsilon}(f_R) = \{f \in \hat{F} \mid \mathbb{E}[L(f)] \leq \mathbb{E}[L(\hat{f}_R)] + \varepsilon\}.
\end{equation}

\noindent The number of models in the Rashomon set is defined as the \textbf{Rashomon Set Size}. It can be used as a metric to measure the severity of the Rashomon effect \cite{Rudin_et_al_2022}. A high value of this metric indicates the presence of many candidate models for the same task, thus a high Rashomon effect. In contrast, a low value suggests a small number of candidate models and a low Rashomon effect. However, this criterion is only calculated based on the number of models in the Rashomon set, making it superficial. It does not reveal the outcome of this effect at the observation or model level. Marx et al. \cite{Marx_et_al_2020} have addressed this by introducing the \textit{ambiguity} and \textit{discrepancy} metrics.\\

\noindent \textbf{Ambiguity.} The ambiguity of a Rashomon Set $R_{L,\varepsilon}(f_R)$ is the ratio of observations in the data $\textbf{X}$ that has conflicting predictions with that reference model $f_R$ and the other models in $R_{L,\varepsilon}(f_R)$:
\begin{equation}
\alpha_{\varepsilon}(f_R) = \frac{1}{n} \sum_{i=1}^{n} \max_{f \in R_{L,\varepsilon}(f_R)} \mathbbm{1}[f(\textbf{x}_i) \neq f_R(\textbf{x}_i)].
\end{equation}

\noindent \textbf{Discrepancy.} The discrepancy of a Rashomon Set $R_{L,\varepsilon}(f_R)$ is the maximum ratio of conflicting observations between the reference model $f_R$ and the other models in $R_{L,\varepsilon}(f_R)$:
\begin{equation}
\delta_{\varepsilon}(f_R) = \max_{f \in R_{L,\varepsilon}(f_R)} \frac{1}{n} \sum_{i=1}^{n} \mathbbm{1}[f(\textbf{x}_i) \neq f_R(\textbf{x}_i)].
\end{equation}

Ambiguity indicates the ratio of observations assigned conflicting predictions by any competing model in the Rashomon set. However, it gives binary information about the conflicting predictions, i.e., it shows a conflict or not and does not characterize the multiplicity of the set in detail. Thus, we propose a new metric called obscurity, which represents the average ratio of conflicting predictions for the observations. It is more useful from a data-centric point of view because it measures the abundance of observation conflict in terms of whether it occurs or not and the average ratio of conflicting predictions. All the metrics take values between zero and one.\\

\noindent \textbf{Obscurity.} The obscurity of a Rashomon Set $R_{L,\varepsilon}(f_R)$ is the average ratios of conflicting predictions for the observations in the data $\textbf{X}$ between the reference model $f_R$ and the other models in $R_{L,\varepsilon}(f_R)$:
\begin{equation}
\gamma_{\varepsilon}(f_R) = \frac{1}{n} \sum_{i=1}^{n} \frac{1}{|R_{L,\varepsilon}(f_R)|} \sum_{f \in R_{L,\varepsilon}(f_R)} \mathbbm{1}[f(\textbf{x}_i) \neq f_R(\textbf{x}_i)].
\end{equation}

\begin{figure}[h!]
    \centering
    \includegraphics[trim=1cm 11cm 0.2cm 6cm, clip, scale = 0.18]{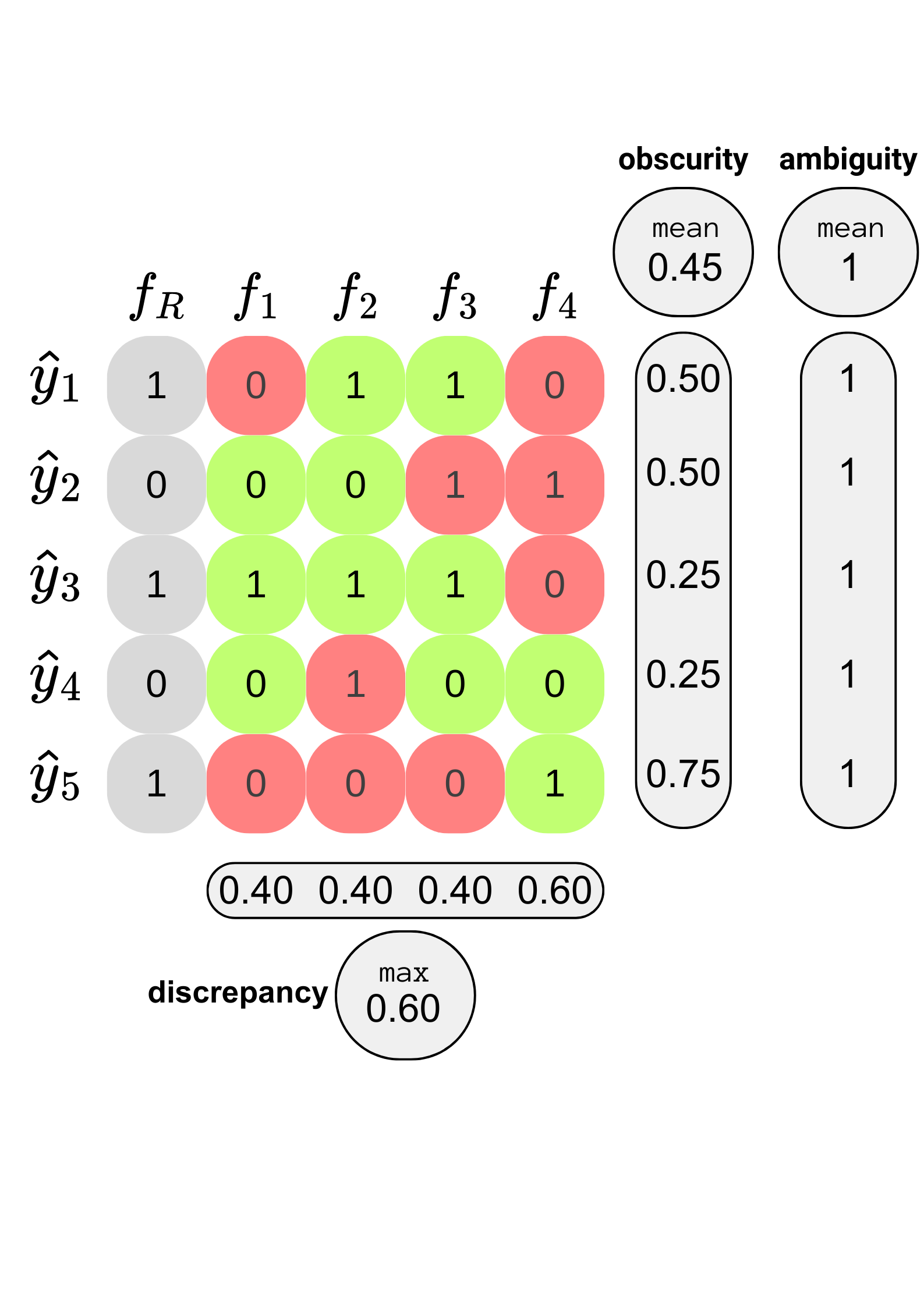}
    \caption{The illustration of relations among \textit{ambiguity}, \textit{discrepancy}, and \textit{obscurity}.}
    \label{fig:amb_disc}
\end{figure}

Fig~\ref{fig:amb_disc} illustrates the computation of \textit{ambiguity}, \textit{discrepancy}, and \textit{obscurity}. Assume that there are five models in the Rashomon set. To simplify the illustration, we analyzed only five samples; however, it is important to note that these two metrics were calculated on all samples. First column represents the reference model predictions $\hat{y}_i = f_R(X_i)$ for observations $i= 1, 2, 3, 4, 5$. The following columns show the models' predictions in the Rashomon set which are $f_1, f_2, f_3$, and $f_4$. The red cells indicate a conflict prediction, and the green cells indicate a consistent prediction of the models in the Rashomon set. The maximum value of the ratio of the conflict predictions in the models is called \textit{discrepancy}, the ratio of the conflict predictions for observations is called \textit{ambiguity}, and the \textit{obscurity} is the mean of the conflict predictions for observations. The computation of these metrics depends on access to models in the Rashomon set, which is computationally infeasible \cite{Hsu_et_al_2024,Donnelly_et_al_2024}. Using specific model classes and retraining models with different hyperparameter setups are the strategies to create the Rashomon set. 

\vspace{-15pt}
\vspace{-15pt}

\section{Experiments}
\label{sec:experiments}

In this section, we conduct experiments to observe the Rashomon effect of balancing methods on model behavior in terms of multiplicity.  We considered four balancing methods: \textit{random oversampling}, \textit{SMOTE} \cite{Chawla_et_al_2002}, \textit{random undersampling}, and \textit{near miss} \cite{Mani_and_Zhang_2003} which are used to improve the prediction performance. The first two of these methods are based on oversampling, and the rest are based on the undersampling of the majority class. In the balancing phase, the train set is balanced using these methods, and the test set remains as it is. We used the \texttt{forester}, a tree-based AutoML tool \cite{Kozak_and_Ruczynski_2023} to create a Rashomon set. We prefer to use this approach because it is more useful than the other ways to access the hypothesis space. It also enables to control of the Rashomon set size using the parameters provided by the Bayesian optimization part of the tool. The Rashomon set is created on the tasks within the imbalanced benchmark dataset proposed by \cite{Stando_et_al_2023}, as provided in Table~\ref{table:benchmark}. It consists of several data from diverse domains. The imbalance ratio (\textit{$\#$samples in majority class/$\#$samples in minority class}) of the datasets varies between $1.54$ and $129.53$. The Rashomon parameter $\varepsilon$ is taken as $0.05$, and the resampling ratio (i.e., imbalanced ratio after balancing) varies as $\{1, 1.05, 1.10, 1.15, 1.20, 1.25\}$. Moreover, the number of optimization rounds \texttt{bayes\_iter} is taken as 5, and the number of trained models \texttt{random\_evals} is taken as 10 in the \texttt{train} function of the \texttt{forester}. 

\vspace{-15pt}
\begin{table}[H]
\centering
\caption{Imbalanced benchmark tabular datasets}
\label{table:benchmark}
\begin{tabular}{lrrr}
\toprule
  \textbf{Dataset} &  \textbf{Imbalanced ratio} &  \textbf{     $\#$Samples} &  \textbf{ 
    $\#$Variables} \\\midrule
          \texttt{spambase} &   1.54 &  4601 &    55 \\
    \texttt{MagicTelescope} &   1.84 & 19020 &    10 \\
\texttt{steel-plates-fault} &   1.88 &  1941 &    13 \\
           \texttt{phoneme} &   2.41 &  5404 &     5 \\
               \texttt{jm1} &   4.17 & 10880 &    17 \\
       \texttt{SpeedDating} &   4.63 &  1048 &    18 \\
               \texttt{kc1} &   5.47 &  2109 &    17 \\
             \texttt{churn} &   6.07 &  5000 &     8 \\
               \texttt{pc4} &   7.19 &  1458 &    12 \\
               \texttt{pc3} &   8.77 &  1563 &    14 \\
           \texttt{abalone} &   9.68 &  4177 &     7 \\
         \texttt{us\_crime} &  12.29 &  1994 &   100 \\
        \texttt{yeast\_ml8} &  12.58 &  2417 &   103 \\
               \texttt{pc1} &  13.40 &  1109 &    17 \\
   \texttt{ozone-level-8hr} &  14.84 &  2534 &    72 \\
              \texttt{wilt} &  17.54 &  4839 &     5 \\
     \texttt{wine\_quality} &  25.77 &  4898 &    11 \\
        \texttt{yeast\_me2} &  28.10 &  1484 &     8 \\
       \texttt{mammography} &  42.01 & 11183 &     6 \\
       \texttt{abalone\_19} & 129.53 &  4177 &     7 \\
\bottomrule
\end{tabular}
\end{table}
\vspace{-10pt}

We utilized permutational variable importance \cite{fisher_et_al_2019} in the second part of the experiments, which focused on measuring the predictive multiplicity and investigating model behavior change. It has been frequently used in explanatory model analysis \cite{Biecek_et_al_2021}. For example, some methods are proposed for determining variable importance through the smoothness of the partial dependence profiles \cite{Greenwell_et_al_2020,Kozak_and_Biecek__2020}. Moreover, some statistics are introduced to measure model behavior change calculating differences between profiles \cite{Stando_et_al_2023,Zhang_et_al_2021,Kobylinska_et_al_2023}. In this study, we propose a new metric which is called \textit{Variable Importance Order Discrepancy} (VIOD) to measure dissimilarity between variable importance orders of the models in the Rashomon set based on Kendall's $\tau$ correlation coefficient \cite{Kendall_1938}: $\zeta_{\varepsilon}(f_R) = \max_{f \in R_{L,\varepsilon}(f_R)} \tau(f_R, f).$ The VIOD of a Rashomon set measures the maximum dissimilarity of variable importance orders between the reference model and the other models in the Rashomon set. It provides the maximum dissimilarity of variable importance orders that could change if the model is replaced with a competing model. We use it to measure the model behavior change regarding permutational variable importance for conducting the model selection responsibly. In addition, we used statistical plots \cite{Patil_2021}, and statistical tests Kruskal-Wallis \cite{Kruskal_and_Wallis_1952}, Friedman \cite{Friedman_1937}, and Dunnett's pairwise test \cite{Dunnett_1955} to summarize and statistically evaluate the findings. The results are given in the following section.

\section{Results}
\label{sec:results}

In this section, the experiments are conducted to explore the research questions considered in the paper.\\

\noindent \textbf{RQ1. How do the balancing methods affect the predictive multiplicity of the models in the Rashomon set?}\\
\noindent Monitoring the impact of balancing methods on the Rashomon set is crucial as multiplicity outlines the potential damage at the individual level induced by the arbitrary model selection \cite{Hsu_and_Calmon_2022}. Thus, we measure the multiplicity of the Rashomon sets after balancing the datasets. 

\begin{figure}[h!]
    \centering
    \includegraphics[scale = 0.3]{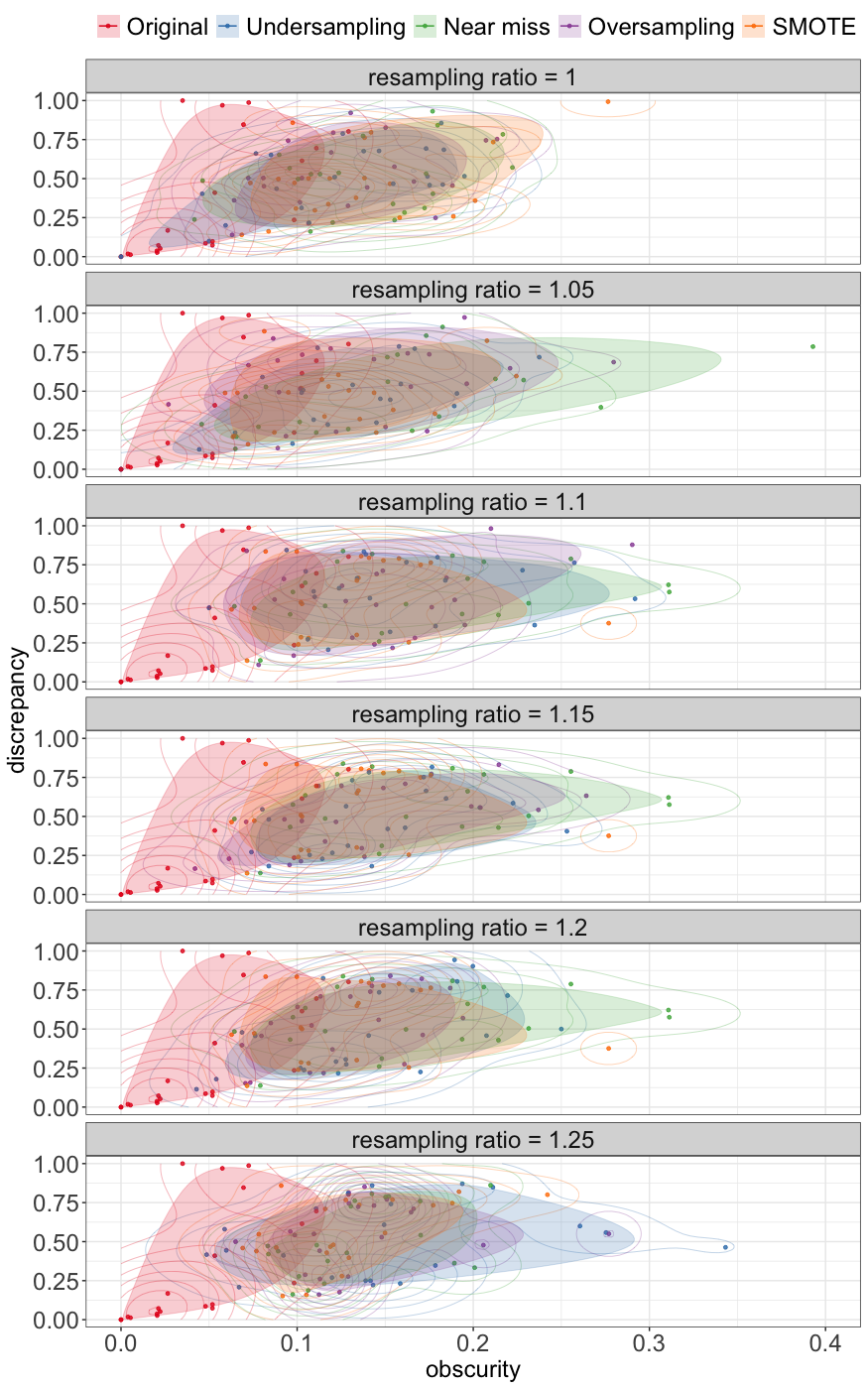}
    \caption{The 2d density plot of the Rashomon metrics \textit{obscurity} and \textit{discrepancy} for different balancing methods and various resampling ratios.}
    \label{fig:rashomon}
\end{figure}

The calculated \textit{obscurity} and \textit{discrepancy} values of the Rashomon sets are given in Figure~\ref{fig:rashomon}, which consists of the individual points representing the Ras\-ho\-mon metrics of the models trained on each dataset listed in Table~\ref{table:benchmark}. The zones show two-dimensional regions where metrics' values are dense. The value of $1.25$ means the frequency of the majority class over that of the minority class. Zones being close to zero on both axes indicate low severity of multiplicity, and moving away from them indicates increasing severity of multiplicity. The zone calculated for the Rashomon set when applied to the balanced dataset using any balancing method, completely overlaps. This allows us to visually interpret that the relevant balancing method does not change the Rashomon metrics of the models. If it does not overlap, it indicates that the balancing method does affect the metrics. Although the zones do not overlap exactly, they are not positioned very differently from each other on the discrepancy axis. However, the zones of the balancing methods on the \textit{obscurity} axis do not overlap with the zones of the models trained on the original dataset. Balancing methods increase the \textit{obscurity} of Rashomon sets.

\begin{figure}[h!]
    \centering
    \includegraphics[scale = 0.35]{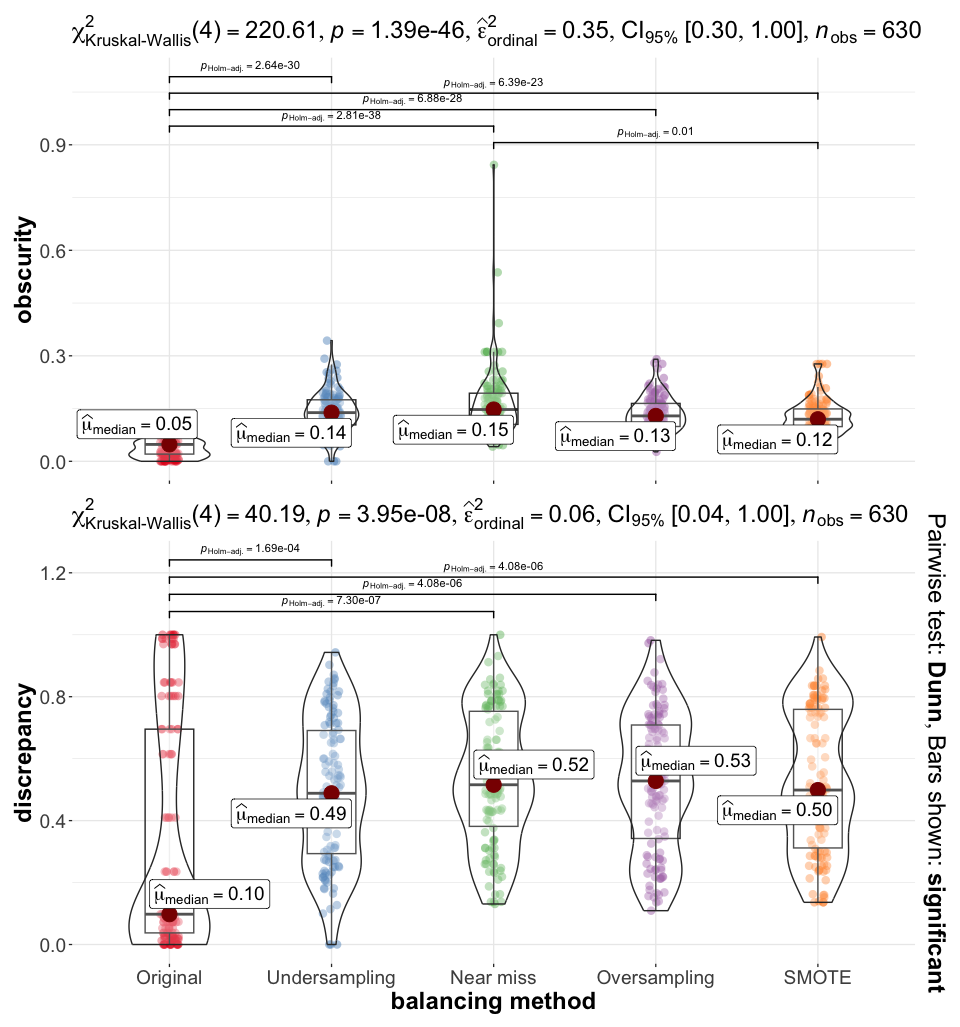}
    \caption{The distribution plots of the Rashomon metrics \textit{obscurity} and \textit{discrepancy} for different balancing methods.}
    \label{fig:comparison_rash_stat}
\end{figure}

In addition to visual examinations of the zones, the effects of balancing methods on Rashomon metrics were compared using the Kruskal-Wallis test, and the results are given in Fig~\ref{fig:comparison_rash_stat}. It consists of the results of the statistical tests Kruskal-Wallis and Dunn's Pairwise tests. The reference bars above each violin indicate statistically significant differences between the medians of groups and the corresponding statistical information. The results show no statistical differences between the Rashomon metrics of the models because the p-value of the tests is lower than 0.05 for both \textit{obscurity} and \textit{discrepancy}. The pairwise test results show the difference's source is the models trained on the original dataset. The Rashomon metrics of the models trained on the original dataset are lower than those trained on the balanced datasets. It is concluded that the balancing methods increase the \textit{discrepancy} and \textit{obscurity} of the Rashomon sets.\\ 

\noindent \textbf{RQ2. How do the balancing methods affect the variable importance order discrepancy of the models in the Rashomon set?}\\
\noindent The models in the Rashomon set may not use similar variables \cite{Poiret_et_al_2023}, and also, the small changes in training data during the training data developments can produce large changes in the outputs \cite{Oh_et_al_2022}. Thus, it is important to check whether the balancing methods inflate multiplicity, and we investigate how they affect model behavior. The order of variable importance is an important way to measure the change in the model behavior. Here, we used the \textit{variable importance order discrepancy} (VIOD) to measure the discrepancy of the variable importance orders in the Rashomon set. Thus, we want to carry out the model selection process responsibly, being aware of how a model selected from the Rashomon set behaves differently from other models in terms of variable importance order. 

\begin{figure}[h!]
    \centering
    \includegraphics[scale = 0.35]{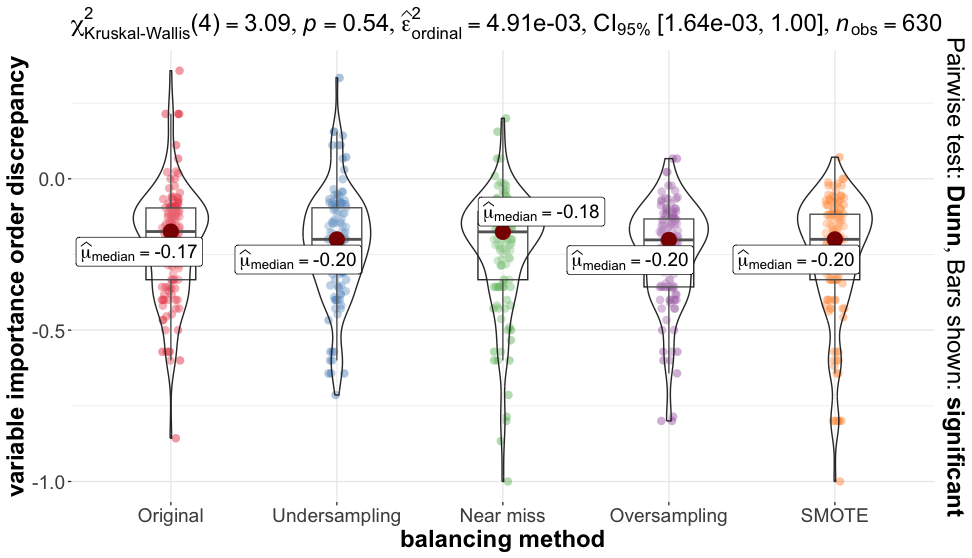}
    \caption{The distribution plots of the Rashomon metric \textit{variable importance order discrepancy} for different balancing methods.}
    \label{fig:comparison_vio}
\end{figure}

The effects of balancing methods on \textit{VIOD} are compared using the Kruskal-Wallis test, and the results are given in Fig~\ref{fig:comparison_vio}. There is no statistically significant difference between the median of the \textit{VIOD} values of the Rashomon set; there is no need to conduct any pairwise comparison test.

\noindent \textbf{RQ3. Can partial resampling be a solution to the model behavior change?}\\
\noindent The partial resampling is proposed to mitigate the bias of balancing methods. Here, we use partial resampling to check whether it can be a solution to mitigate the effect of balancing methods on the model behavior. 

\begin{figure}[h!]
    \centering
    \includegraphics[scale = 0.4]{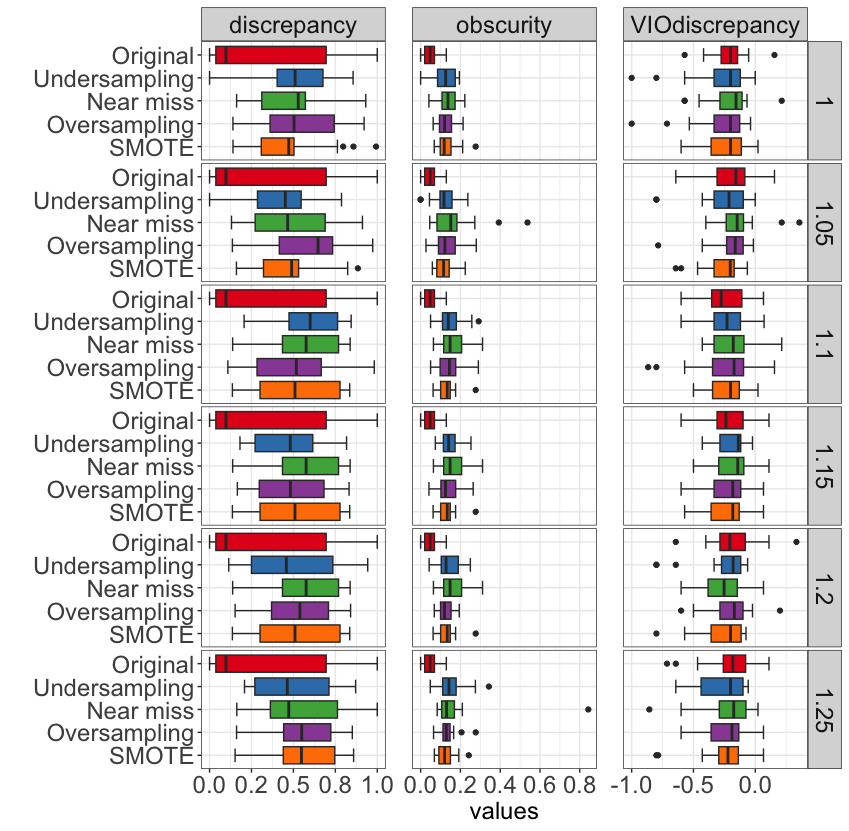}
    \caption{The distribution plots of the Rashomon metric \textit{variable importance order discrepancy} for different balancing methods and varying partial resampling ratios.}
    \label{fig:partial_resampling}
\end{figure}

The distribution of the Rashomon metrics \textit{obscurity}, \textit{discrepancy}, and \textit{variable importance order discrepancy} is given for balanced and partially balanced datasets in Fig~\ref{fig:partial_resampling}. No pattern between the metrics over resampling ratios is seen. We conduct the Friedman test to test the effect of resampling ratios on the Rashomon metrics. The comparison of the Rashomon metrics over the balancing methods under the block effect of resampling ratios was performed individually using Friedman’s test, which showed statistically significant differences. The results are as follows: $\chi^2_{(5)} = 6.8286$, $p = 0.2337$, $\chi^2_{(5)} = 9$, $p = 0.1091$, and $\chi^2_{(5)} = 3.7429$, $p = 0.5870$, respectively. There is no statistically significant difference between the Rashomon metrics and resampling ratios. Thus, partial resampling is not a solution for the multiplicity problem of balancing methods.\\

\begin{figure}[h!]
    \centering
    \includegraphics[scale = 0.338]{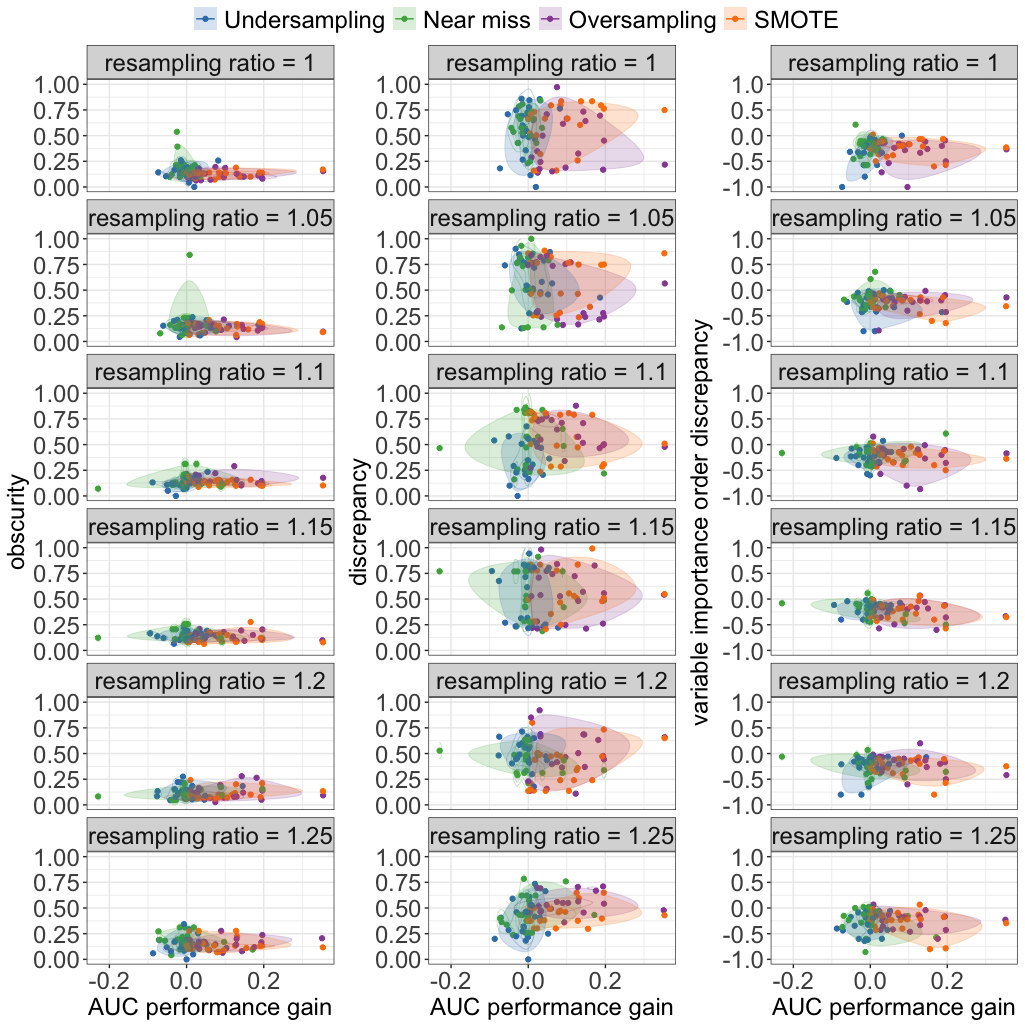}
    \caption{The performance gain plots of \textit{obscurity}, \textit{discrepancy}, \textit{variable importance order discrepancy} for different balancing methods and varying partial resampling ratios.}
    \label{fig:performance_gain}
\end{figure}

\noindent \textbf{RQ4. Can performance gain plot be a method to monitor the trade-off between performance gain and multiplicity?}\\
We considered alternative solutions after determining that the partial resampling method was insufficient to combat the model behavior change. One of the potential solutions is the \textit{performance-gain plot}, which is proposed to monitor the trade-off between the effect of balancing methods and performance gain \cite{Stando_et_al_2023}. Here, we expand it for Rashomon metrics to monitor the multiplicity in the Rashomon set over performance gain. The performance-gain plot for Rashomon metrics is given in Fig~\ref{fig:performance_gain}. The horizontal axis shows the performance gain in terms of AUC, which is calculated by the difference between the AUC of the original data and the balanced data. The zero indicates no gain, and the negative values indicate performance loss. The vertical axes are limited between zero and one for \textit{obscurity} and \textit{discrepancy}, but it is between minus one and one for \textit{variable importance order discrepancy}. 

Moving the zones towards the positive way on the horizontal axis indicates an increase in performance gain, and moving towards the negative way on the vertical axis indicates a decrease in the multiplicity. The oversampling-based resampling methods \textit{Oversampling} and \textit{SMOTE} improve the performance higher than others in all cases. One of the interesting findings is that although the performance gain increases when the data is more balanced with the \textit{Near miss} method, the obscurity and discrepancy also increase. However, it should be noted that the performance gain here is not noteworthy. Significantly, \textit{Undersampling} and \textit{Near miss} methods can not improve the prediction performance in some datasets. This is a known deficiency of the balancing methods \cite{Elor_et_al_2022}. Although the \textit{Oversampling} and \textit{SMOTE} methods seem to provide close performance gains, it can be said that the \textit{Oversampling} method leads Rashomon sets consisting of models with similar variable importance order.

\section{Conclusions}
This paper investigated the effect of one commonly used data-centric AI approach in the training data development phase, balancing methods on predictive multiplicity by addressing different \textbf{RQs}. It is essential because exploring the Rashomon set provides some advantages and challenges. It can be utilized to select models that satisfy additional properties without compromising accuracy, such as fairness, interpretability, and stability \cite{Paes_et_al_2023}. However, it may be problematic because the models in the Rashomon set rely on different variables, and the blind selection of the model can produce more profitable or unfavorable predictions to the individuals \cite{Meyer_et_al_2023}.  

We conduct experiments on benchmark datasets with varying imbalanced ratios. For the \textbf{RQ1. How do the balancing methods affect the predictive multiplicity of the models in the Rashomon set?} We found that balancing methods increase Rashomon metrics. In the \textbf{RQ2. How do the balancing methods affect the variable importance order discrepancy of the models in the Rashomon set?} We investigated the effect of balancing methods with a new metric we proposed on the model behavior through variable importance order and did not observe a statistically significant order change that contradicts the findings of Alarab and Prakoonwit \cite{Alarab_and_Prakoonwit_2022}. However, considering that they conducted their experiments on only two data sets, we think the results we obtained on a larger number of datasets are more comprehensive and reliable. The results we obtained above show that the balancing methods inflate the Rashomon metrics, making it necessary to develop solutions. First, in the \textbf{RQ3. Can partial resampling be a solution to the model behavior change?} We examined whether the partial resampling approach suggested by Kamalov et al. \cite{Kamalov_et_al_2022} could be a solution. However, we found that partial resampling does not work. Then, focusing on \textbf{RQ4. Can the extended version of the performance gain plot be a solution to monitor the trade-off between performance gain and multiplicity?} and expanding the performance-gain plot and monitoring the balance between performance and Rashomon metrics, we showed that choosing the method that best suits the data could be a solution. Its easy-to-use structure makes the performance-gain plot flexible and expandable across the intended Rashomon metrics.

In conclusion, it is vital to consider the recent discussions on balancing methods during the training data development phase of data-centric AI, incorporating an additional dimension. Our experiments show that balancing methods increase the predictive multiplicity and the risks involved in the model selection process from the Rashomon set. Therefore, researchers should consider the risk of multiplicity when using balancing methods, which are a haven in imbalanced classification problems. Using the extended performance-gain plot for the Rashomon effect is essential to monitor this issue and conduct the modeling process responsibly.

\section*{Further Research}

\noindent There are still opportunities for further exploration in this field. To mitigate the predictive multiplicity, identifying the source of the Rashomon effect is a hot topic \cite{Meyer_et_al_2023}. First, it is stated that oversampling methods increase data complexity \cite{Komorniczak_et_al_2022}. The question of whether there is a connection between data complexity and the Rashomon effect can be investigated, e.g., ``Does the increase in data complexity due to oversampling methods lead to an increase in the Rashomon effect?". Second, similar to the study in \cite{Junior_and_Pisani_2022}, it can be investigated whether cost-sensitive methods, such as resampling methods, have a lesser impact on the Rashomon effect. Third, special cases of imbalance situations, such as class overlap and small disjunct, as mentioned in \cite{Garcia_et_al_2012,Prati_et_al_2015}, can also be examined.
\subsubsection*{Acknowledgment.} This study is funded by the Scientific and Technological Research Council of Turkiye (TÜBİTAK) through the 2224-A Grant Program for Participation in Scientific Meetings Abroad (application no. 1919B022401776) and Eskisehir Technical University Scientific Research Projects Commission under grant no. 24ADP116.

\subsubsection*{Supplemental Materials.}

The materials for reproducing the experiments and the benchmark datasets are available in the repository: \url{https://github.com/mcavs/ECML2024_Imbalanced_Rashomon_Paper}.


\end{document}